\documentclass[11pt]{article}

\usepackage[final]{acl}

\usepackage{times}
\usepackage{latexsym}

\usepackage[T1]{fontenc}

\usepackage[utf8]{inputenc}

\usepackage{microtype}

\usepackage{inconsolata}

\usepackage{graphicx}
\usepackage{amsmath}
\usepackage{amsfonts}
\allowdisplaybreaks
\usepackage{amsthm}
\newtheorem{definition}{Definition}
\newtheorem{theorem}{Theorem}
\newtheorem{proposition}{Proposition}
\newenvironment{proofsketch}{\noindent{\it Proof Sketch}\hspace*{1em}}{\qed\medskip}
\usepackage{hyperref}

\title{LayerNorm Induces Recency Bias in Transformer Decoders}
    
\author{
Junu Kim$^{1, 2}$\thanks{Work done during an internship at Microsoft Research Asia.} Xiao Liu$^{2}$  Zhenghao Lin$^{2}$  Lei Ji$^{2}$  Yeyun Gong$^{2}$  Edward Choi$^{1}$ \\
$^{1}$ KAIST $^{2}$ Microsoft Research \\
\texttt{\{kjune0322,edwardchoi\}@kaist.ac.kr} \\
\texttt{\{xiao.liu.msrasia,zhenghaolin,leiji,yegong\}@microsoft.com} \\
}

\begin{document}
\maketitle
\begin{abstract}
Causal self-attention provides positional information to Transformer decoders.
Prior work has shown that stacks of causal self-attention layers alone induce a positional bias in attention scores toward earlier tokens.
However, this differs from the bias toward later tokens typically observed in Transformer decoders, known as recency bias.
We address this discrepancy by analyzing the interaction between causal self-attention and other architectural components.
We show that stacked causal self-attention layers combined with LayerNorm induce recency bias.
Furthermore, we examine the effects of residual connections and the distribution of input token embeddings on this bias.
Our results provide new theoretical insights into how positional information interacts with architectural components and suggest directions for improving positional encoding strategies.
\end{abstract}

\section{Introduction}

In sequence modeling with Transformer decoders \citep{vaswani2017attention}, the way positional information is provided to the model is closely tied to performance \citep{dufter2022position} and its ability to generalize to longer sequence lengths \citep{zhao2024length}.
Among the components of a Transformer decoder layer, positional encodings and the causal mask are responsible for supplying positional information \citep{haviv2022transformer, kazemnejad2023impact, chi2023latent}.
While the mechanisms by which positional encodings provide positional information have been extensively studied \citep{barberoround, su2024roformer, presstrain}, the corresponding process for the causal mask remains less well understood.

Recent research shows that simply stacking causal self-attention layers can induce an attention bias toward earlier tokens, thereby providing positional information \citep{wuemergence}.
However, empirical studies of full Transformer decoder layers yield contrasting results.
Specifically, \citet{zuo2025position} show that Transformer decoders exhibit an attention bias toward more recent tokens rather than earlier ones.
This phenomenon, known as recency bias, is characteristic of many positional encoding methods \citep{su2024roformer, presstrain, vaswani2017attention}.
The discrepancy between these findings suggests that additional architectural components, such as LayerNorm \citep{ba2016layer} or residual connections \citep{he2016deep}, may modulate the positional information induced by the causal mask.

By examining the effects of other architectural components, we show that LayerNorm induces recency bias in Transformer decoders without positional encoding.
Formally, stacking causal self-attention layers with LayerNorm induce recency bias, consistent with the observations of \citet{zuo2025position}.
We further analyze the effects of residual connections and the distribution of input token embeddings on recency bias, both quantitatively and qualitatively.
Together, these findings provide theoretical insights into improving positional encoding and length generalization in Transformer decoders.

\section{Theoretical Analysis}

\subsection{Preliminaries}

{ 
\setlength{\abovedisplayskip}{4pt}
\setlength{\belowdisplayskip}{4pt}
\setlength{\abovedisplayshortskip}{0pt}
\setlength{\belowdisplayshortskip}{0pt}

Formally, a single-head pre-LayerNorm \citep{xiong2020layer} Transformer decoder layer defines a function 
$f^{(l)}:\mathbb{R}^{n\times d} \to \mathbb{R}^{n\times d}$ with 
$X^{(l)} = f^{(l)}(X^{(l-1)})$, 
where $X^{(l)}$ denotes the output of the $l$-th Transformer decoder layer, and $n$ and $d$ correspond to the number of input tokens and the model hidden size, respectively.
The superscripts indicate layer indices, and $X^{(0)}$ corresponds to the input token embeddings. We omit the superscripts when clear from context.

First, the normalized input $Y^{(l)}$ is produced by the $\mathrm{LayerNorm}$ operation:
\begin{equation*}
Y^{(l)} = \mathrm{LayerNorm}(X^{(l-1)}).
\end{equation*}

Then, the query, key, and value matrices $Q$, $K$, and $V$ are computed using the learnable weight matrices $W_Q, W_K, W_V \in \mathbb{R}^{d \times d}$:
\begin{gather*}
Q = Y W_Q,\quad K = Y W_K,\quad V = Y W_V.
\end{gather*}

The attention score matrix $S$ is computed as:
\begin{gather*}
S = \mathrm{Causal}\left(\frac{QK^\top}{\sqrt{d}}\right),
\end{gather*}
where $\mathrm{Causal}(\cdot)$ applies a strictly upper-triangular causal mask (setting masked entries to $-\infty$) to prevent attention to future positions.

Thus, the attention weights $A$ are defined by the row-wise softmax operation:
\begin{gather*}
A = \mathrm{Softmax}(S).
\end{gather*}

The attention output $O^{(l)}$ is then computed using a learnable output projection matrix $W_O \in \mathbb{R}^{d \times d}$ together with a residual connection:
\begin{gather*}
O^{(l)} = (A V) W_O + X^{(l-1)}.
\end{gather*}

Finally, the hidden state $X^{(l)}$ is computed as:
\begin{gather*}
X^{(l)} = \mathrm{FFN}(\mathrm{LayerNorm}(O^{(l)})) + O^{(l)},
\end{gather*}
where $\mathrm{FFN}$ denotes a position-wise feed-forward network, typically consisting of two linear layers separated by a nonlinear activation function.

Here, we formally define recency bias as the property that the attention score assigns a higher score to a closer key than to a more distant key for a fixed query.
\begin{definition}
    The attention score $S$ exhibits recency bias if $S_{ij} > S_{ik}$ for all query indices $i$ and key indices $j$ and $k$ satisfying $i \geq j > k$.
\end{definition}

}

\subsection{LayerNorm}\label{math}
Here, we show that stacked causal self-attention layers with LayerNorm can induce a recency bias, even when the input sequence has no causal dependency, no learnable parameters, and no feed-forward modules.
We begin with the case where the input token embeddings follow a normal distribution with zero mean and variance $1/d$, following \citet{reddymechanistic} and \citet{wuemergence}.
We first ignore the residual connection, and then consider cases with residual connections or non-normal input distributions in the following subsections.
\begin{theorem}\label{thm}
    Let input token embeddings follow $\mathcal{N}(0, \mathbb{I}_d/d)$, and let the architecture be composed of stacked LayerNorm and causal self-attention layers. For hidden sizes $d \gg 1$, the attention score of the second layer $S^{(2)}$ exhibits a recency bias.
\end{theorem}

Here, we provide a sketch of the proof; the full version  can be found in Appendix \ref{proof_thm}.

\begin{proofsketch}
    Following the assumption, each input token embedding satisfies $x^{(0)}_i \sim \mathcal{N}(0, \mathbb{I}_d/d)$. 
    Under the stated simplifications, a single Layer $f(X)$ acts as: 
    \begin{align}
        f(X) = \mathrm{Softmax}(\mathrm{Causal}(YY^\top / \sqrt{d}))Y,\\
        Y= \mathrm{LayerNorm}(X) \label{eq:simplified}.
    \end{align}
    The operator $\mathrm{Causal}(\cdot)$ applies a strictly upper triangular mask so that a query at position $i$ attends only to keys at position $j \leq i$, and the softmax is applied row-wise:
    \begin{equation}
    \begin{aligned}[b]
        A^{(1)}_{ij} =
        \begin{cases}
            \frac{e^{\langle y_i, y_i\rangle/\sqrt{d}}}{e^{\langle y_i, y_i\rangle /\sqrt{d}} + \sum_{k=1}^{i-1} e^{\langle y_i, y_k\rangle/\sqrt{d}} }& (i=j) \\
            \frac{e^{\langle y_i, y_j\rangle /\sqrt{d}}}{e^{\langle y_i, y_i\rangle /\sqrt{d}} + \sum_{k=1}^{i-1} e^{\langle y_i, y_k\rangle /\sqrt{d}} } & (i>j) \\
            0 & (i<j)
        \end{cases}.
    \end{aligned} \label{eq:softmax_causal}
    \end{equation}
    Since $d \gg 1$, we can apply the law of large numbers.
    For large $d$, LayerNorm can be approximated as $\mathrm{LayerNorm}(X) \approx \sqrt{d}X/ ||X||_2 $.
    Consequently, $\langle y^{(1)}_i, y^{(1)}_i \rangle / \sqrt{d} = \sqrt{d}$.
    For $i \ne j$, the cross term $\langle y_i^{(1)}, y_j^{(1)} \rangle / \sqrt{d}$ converges in distribution to $\mathcal{N}(0,1)$, which is negligible compared to $\sqrt{d}$. We therefore approximate these terms by zero.
    Under this approximation,
    \begin{equation}
    \begin{aligned}[b]
        x^{(1)}_i = \sum_{j=1}^{i} A^{(1)}_{ij} y^{(1)}_j 
        = \frac{
        e^{\sqrt{d}} y_i^{(1)} + \sum_{k=1}^{i-1} y^{(1)}_k
        }{e^{\sqrt{d}}+ i-1}.
    \end{aligned}
    \end{equation}
    Since $x_i^{(1)}$ is a linear combination of zero-mean vectors, the same LayerNorm approximation applies at the next layer. 
    Writing $S^{(2)}_{ij}$ in terms of $x^{(1)}$ and applying the law of large numbers again, for $i>j$ we obtain
    \begin{equation}
    \begin{aligned}[b]
        S^{(2)}_{ij} &= \frac{\langle y_i^{(2)}, y_j^{(2)} \rangle}{\sqrt{d}} 
        =\frac{d\langle x_i^{(1)}, x_j^{(1)} \rangle}{\sqrt{d}|| x_i^{(1)}|| \cdot || x_j^{(1)} ||} \\
        &= \frac{\sqrt{d}(e^{\sqrt{d}} + j - 1)}{\sqrt{e^{2\sqrt{d}} + i - 1}\sqrt{e^{2\sqrt{d}}+j-1}}.
    \end{aligned}\label{eq:s_ij_base}
    \end{equation}

    For fixed $i$, this expression is strictly increasing in $j$. For the diagonal case, $S^{(2)}_{ii} = \sqrt{d}$ by construction, and clearly $S^{(2)}_{ii} > S^{(2)}_{i,i-1}$. 
    Therefore, $S^{(2)}_{ij} > S_{ik}^{(2)}$ for all $i \geq j > k$, exhibiting recency bias.
\end{proofsketch}

\begin{figure*}[!t]
    \centering
    \includegraphics[width=\linewidth, clip]{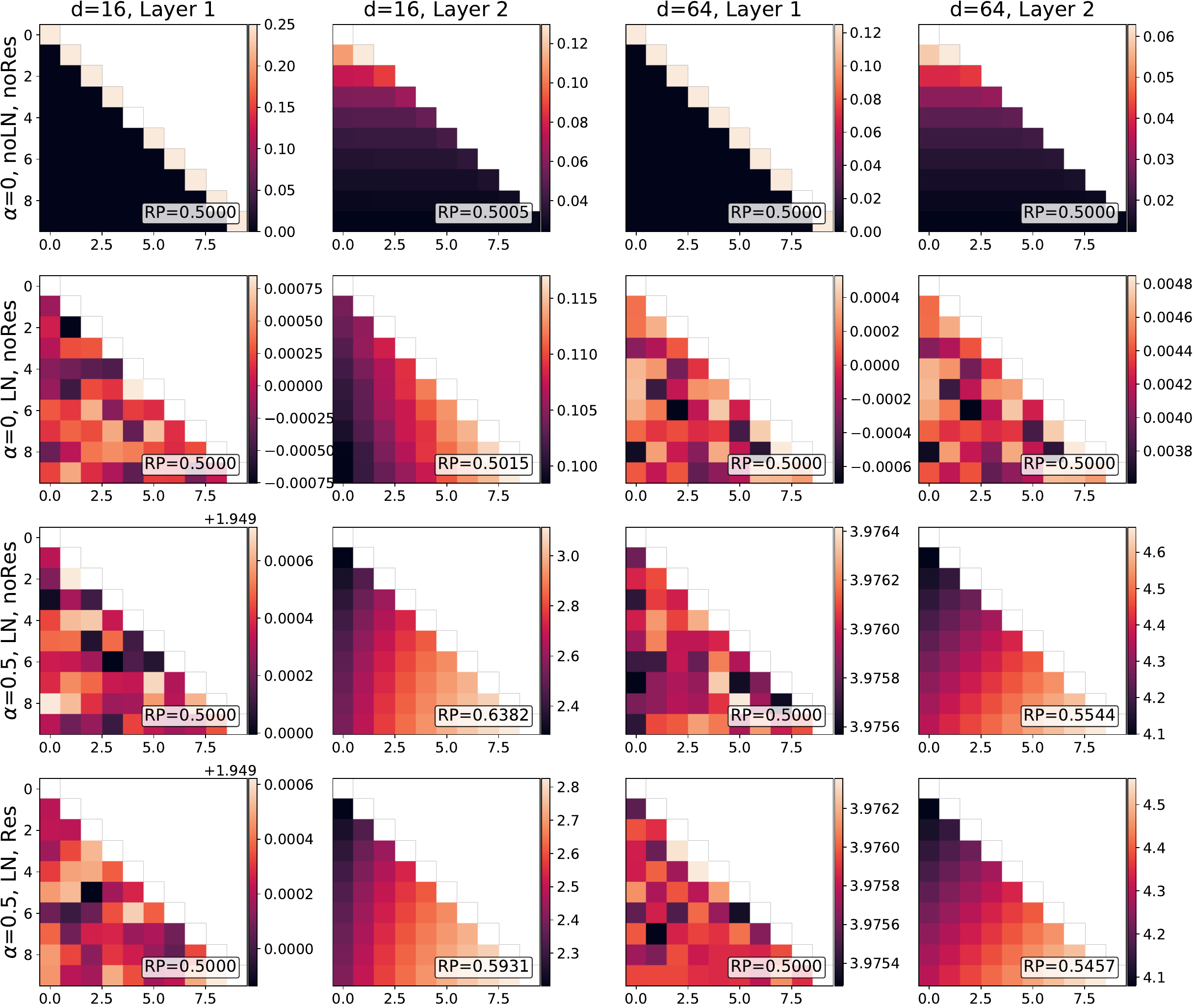}
    \caption{Visualization of the attention scores using a simulation with $d=16$ (left) and $d=64$ (right) for layers 1 and 2. LN and Res correspond to LayerNorm and residual connections, respectively.
    The y-axis represents query indices, and the x-axis represents key indices.
    To clearly visualize the recency bias, we masked the diagonal elements except those in the first row.
    Results for layers 3 and 4 are provided in Figure~\ref{fig:extended_layers}, and results for $d=256$ and $d=1024$ are provided in Figure~\ref{fig:extended_dims}.}
    \label{fig:l2norm}
\end{figure*}

\begin{proposition}\label{nol2}
Without LayerNorm, $S^{(2)}$ does not exhibit recency bias.
\end{proposition}
The proof follows a similar approach to Theorem \ref{thm} and can be found in Appendix \ref{proof_nol2}.
Note that this result is consistent with \citet{wuemergence}, confirming that simply stacking self-attention layers does not induce recency bias.

To summarize, under the minimal assumptions that the input token embeddings follow a normal distribution and that $d \gg 1$, we show that LayerNorm induces a recency bias at $S^{(2)}$.
We empirically show the behavior of later layers in Section \ref{simul}.

\subsection{Residual Connection}
In addition, we evaluate the effect of the residual connection on recency bias.
\begin{proposition}\label{residual}
    Regardless of the existence of the residual connection, $S^{(2)}$ has recency bias.
\end{proposition}
The proof can be found in Appendix \ref{apd:anisotropic}.
We empirically demonstrate its effect on the causal bias in Section \ref{simul}.

\subsection{Distribution of Input Token Embeddings}

While we assume that the input token embeddings follow a normal distribution in the previous sections, the input embeddings of pre-trained Transformer decoder models are typically anisotropic; each token embedding has a high cosine similarity with the others \citep{ethayarajh-2019-contextual}.
To evaluate the effect of anisotropy in input token embeddings on recency bias, we model anisotropic embeddings $x_i^{(0)}$ with anisotropy level $\alpha$ by adding a shared vector $v$ to independent Gaussian noise $\epsilon_i$, scaled by a factor $\sqrt{\frac{\alpha}{1-\alpha}}$. Specifically,

\begin{align}
    x_i^{(0)} = \epsilon_i + \sqrt{\frac{\alpha}{1 - \alpha}}\, v,
    \label{eq:dist}
\end{align}
where $\epsilon_i$ and $v$ are independently drawn from $\mathcal{N}(0, \mathbb{I}_d / d)$.

\begin{proposition}\label{anisotropic}
Regardless of the anisotropy of the input token embeddings, $S^{(2)}$ exhibits recency bias.
\end{proposition}

The proof can be found in Appendix \ref{apd:anisotropic}.
We analyze the effect of anisotropic input embeddings in the following section.

\section{Empirical Analysis}\label{simul}

We further examine how recency bias is induced using a simulation of a Transformer without learnable parameters or positional encodings, as defined in Eq. \ref{eq:simplified}.
Specifically, we sample 10 vectors of dimension $d \in \{16, 64\}$ drawn from the distribution defined in Eq. \ref{eq:dist}.
Figure \ref{fig:l2norm} shows the simulated attention scores for layers 1 and 2, with $d=16$ on the left and $d=64$ on the right, averaged over 10{,}000,000 simulations.\footnote{The source code is publicly available at: \url{https://github.com/starmpcc/layernorm_recency_bias}.}

To quantitatively assess the presence of recency bias, we introduce the recency probability (RP) metric, inspired by the adjacency probability of \citet{zuo2025position}.
Given an attention score matrix $S_{ij}$, the RP is defined as the probability that $S_{ij} > S_{ik}$ for any $i > j > k$, excluding diagonal entries.
Under this definition, if the average RP is significantly larger than $0.5$, this indicates the presence of a recency bias.

First, consider the case without LayerNorm, corresponding to the first row of Figure \ref{fig:l2norm}.
As expected, no attention bias is observed in the first layer.
In the second layer, the attention scores in each row are uniform across all positions except for the diagonal elements, for both $d=16$ and $d=64$.
The RP values remain close to $0.5$ across all conditions, confirming the absence of recency bias.
This behavior is consistent with Proposition \ref{nol2} and Figure 2 of \citet{wuemergence}, which show that stacking causal self-attention layers alone does not induce recency bias, but instead leads to a bias toward earlier tokens.

The second row of Figure \ref{fig:l2norm} corresponds to the case with LayerNorm and $\alpha = 0$.
For $d=64$, the attention scores for the off-diagonal elements are nearly uniform, and no clear recency bias is observed ($\mathrm{RP} = 0.5000$).
This behavior can be explained by Eq.~\ref{eq:s_ij_base}: the $e^{\sqrt{d}}$ term dominates the contribution of $j$, effectively suppressing the recency bias.
However, when $d=16$, a mild recency bias emerges even with $\alpha=0$ ($\mathrm{RP} = 0.5015$), because the exponential term in the denominator is less dominant at smaller dimensions.

\begin{figure}
    \centering
    \includegraphics[width=\linewidth, trim={0 15 0 10}, clip]{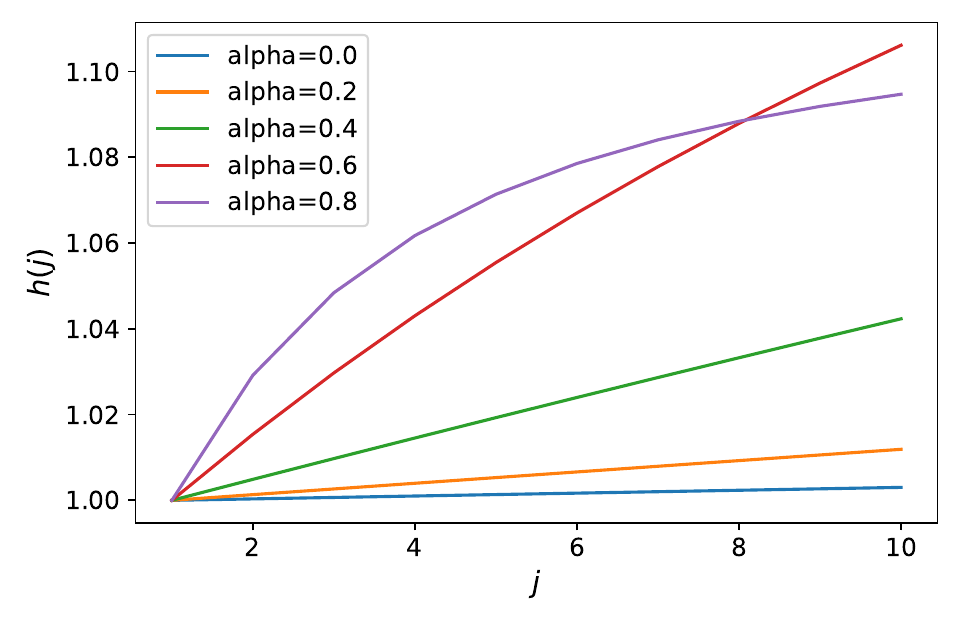}
    \caption{Visualization of $h(j)$ over key index $j$, for multiple values of $\alpha$.}
    \label{fig:alpha}
    \vspace{-0.5em}
\end{figure}

The third row of Figure~\ref{fig:l2norm} corresponds to the anisotropic input distribution with $\alpha = 0.5$.
In the second layer, the attention scores strictly increase as $j$ increases for a fixed $i$, indicating a clear recency bias for both $d=16$ ($\mathrm{RP} = 0.6382$) and $d=64$ ($\mathrm{RP} = 0.5544$).
This behavior can be explained by examining the formulation of $S^{(2)}$.
Equation~\ref{eq:s_ij_base} can be decomposed into a term that depends on $i$ and a term that depends on $j$.
Even when $\alpha \neq 0$, a similar decomposition holds, allowing us to write $S_{ij}^{(2)} = g(i)h(j)$.
For each row, $i$ is fixed, so $g(i)$ can be treated as a constant, and the variation in the attention scores is governed solely by $h(j)$.
The explicit form of $h(j)$ is given in Appendix Eq.~\ref{eq:hj}, and its dependence on $\alpha$ is illustrated in Fig.~\ref{fig:alpha}.
When $\alpha = 0$, the $e^{\sqrt{d}}$ term dominates, causing $h(j)$ to grow slowly.
In contrast, when $\alpha \neq 0$, $h(j)$ increases rapidly, which is consistent with the observed results.

This result is also consistent with \citet{zuo2025position}, which show that hidden states associated with nearby query–key pairs exhibit high cosine similarity in Transformer decoders without positional encoding.
Since cosine similarity is equivalent to the inner product after $\ell_2$ normalization, and $\ell_2$ normalization can in turn be approximated by LayerNorm with a scaling factor of $\sqrt{d}$ under mild assumptions, the observed phenomenon can be explained by our theory.

The last row of Figure \ref{fig:l2norm} corresponds to the case with a residual connection and $\alpha = 0.5$.
Compared to row 3, the recency bias observed in row 4 is less pronounced, as reflected by lower RP values ($\mathrm{RP} = 0.5931$ for $d=16$ and $\mathrm{RP} = 0.5457$ for $d=64$ in layer 2).
Adding a residual connection adds $x_i^{(0)}$ to $x_i^{(1)}$.
Since some components are not shared between $x_i^{(0)}$ and $x_j^{(0)}$, the proportion of components shared by $x_i^{(1)}$ and $x_j^{(1)}$ decreases.
This reduces the overall scale of the off-diagonal attention scores, making the recency bias less pronounced.
A similar analysis for Figure~\ref{fig:alpha} with residual connections (Figure~\ref{fig:alpha_resid}), along with attention score visualizations for different values of~$\alpha$ (Figures~\ref{fig:extended_alpha} and~\ref{fig:extended_alpha_resid}), is provided in Appendix~\ref{apd:res}.

\section{Conclusion}
In this work, we show that causal self-attention, when combined with LayerNorm, induces a recency bias in attention scores.
We further investigate the effects of residual connections and anisotropic input embeddings on this bias.
Our theoretical framework naturally extends to modern attention variants used in contemporary LLMs, including Multi-Head Attention (MHA) \citep{vaswani2017attention}, Multi-Query Attention (MQA) \citep{shazeer2019fast}, and Grouped-Query Attention (GQA) \citep{ainslie2023gqa}.
In these cases, the $\sqrt{d}$ term in the derivation is replaced by $\sqrt{d_h}$, where $d_h$ is the per-head dimension, and all dimension-dependent terms are correspondingly replaced by $d_h$, preserving the structure of the derivation and leading to the same qualitative conclusions.

Importantly, although LayerNorm can induce a recency bias, this does not imply relativity.
Relativity in positional encoding refers to modeling positional information based on pairwise token distances \citep{shaw2018self}, but the attention scores induced by causal masking and LayerNorm do not satisfy this property.
The recency bias in our analysis depends on both the query and key positions rather than solely on their relative distance, meaning that positional interactions between earlier and later tokens may be non-uniform within a single sequence.
This non-uniform bias may adversely affect the length generalization capability of Transformer decoders. Therefore, exploring mitigation strategies, such as modifying the causal mask to counteract this uneven bias, appears promising.
This suggests that the resulting behavior may differ from that of typical relative positional encodings, including RoPE \citep{su2024roformer} and ALiBI \citep{presstrain}.
We believe these findings provide valuable insights for the design of future positional encoding methods and for research on length generalization.

\section*{Limitations}
This study has several limitations.
We do not analyze the effects of feed-forward networks, other learnable parameters, or multi-head attention within Transformer decoder layers.
Additionally, the relationship between recency bias and overall model performance is not evaluated.
Finally, although most modern Transformer decoder-based models use RoPE \citep{su2024roformer} for positional encoding, its interaction with the positional information induced by causal self-attention has not been examined.

\section*{Acknowledgments}
This work was supported by Microsoft Research Asia, the Institute of Information \& Communications Technology Planning \& Evaluation (IITP) grants (No.RS-2019-II190075, No.RS-2022-II220984, No.RS-2024-00436680), and National Research Foundation of Korea (NRF) grant (RS-2026-25484088), funded by the Korea government (MSIT).

\bibliography{custom}
\newpage
\appendix
\onecolumn
\section{Proofs} \label{apd:deriv}
\subsection{Proof of Theorem \ref{thm}}\label{proof_thm}
We additionally present the skipped derivation of Equation \ref{eq:s_ij_base}, starting from $S_{ij}^{(2)}=\frac{d\langle x_i^{(1)}, x_j^{(1)} \rangle}{\sqrt{d}|| x_i^{(1)}|| \cdot || x_j^{(1)} ||}$.
\begin{proof}
    \begin{gather}
        \big(\langle x_i^{(1)}, x_j^{(1)} \rangle \big)_{i>j}
        = \frac{(e^{\sqrt{d}} y_i^{(1)} + \sum_{k=1}^{i-1} y^{(1)}_k)(e^{\sqrt{d}} y_j^{(1)} + \sum_{l=1}^{j-1} y^{(1)}_l)
        }{(e^{\sqrt{d}}+ i-1)(e^{\sqrt{d}}+ j-1)}\\
        = \frac{e^{2\sqrt{d}} \langle y_i^{(1)}, y_j^{(1)} \rangle + e^{\sqrt{d}}\sum_{l=1}^{j-1} \langle y^{(1)}_i, y^{(1)}_l \rangle + e^{\sqrt{d}} \sum_{k=1}^{i-1} \langle y^{(1)}_j, y^{(1)}_k \rangle + \sum_{k=1}^{i-1} \sum_{l=1}^{j-1} \langle y^{(1)}_k y^{(1)}_l \rangle}
        {(e^{\sqrt{d}}+ i-1)(e^{\sqrt{d}}+ j-1)}.
        \label{eq:x2x2}
    \end{gather}
    By the law of large numbers, $\langle y^{(1)}_i, y^{(1)}_j \rangle = 0$ for $i \neq j$, and $\langle y^{(1)}_i, y^{(1)}_i \rangle = d$. 
    Thus,
    \begin{align}
        \big( \langle x_i^{(1)}, x_j^{(1)} \rangle \big)_{i>j}
        &= \frac{d(e^{\sqrt{d}} + j-1)}{(e^{\sqrt{d}}+ i-1)(e^{\sqrt{d}}+ j-1)} .
    \end{align}
    For the $||x_i^{(1)}||$ term in the denominator, we again apply the law of large numbers to the case $i=j$ in Eq. \ref{eq:x2x2}:
    \begin{align}
        ||x^{(1)}_i||_2^2 = \langle x^{(1)}_i, x^{(1)}_i \rangle 
        = \frac{d(e^{2\sqrt{d}} + i-1)}
        {(e^{\sqrt{d}} + i -1)^2}.
    \end{align}
    Therefore, for $i>j$, 
    \begin{align}
        S_{ij}^{(2)} = \frac{\langle y^{(2)}_i, y^{(2)}_j \rangle}{\sqrt{d}}
        = \frac{d\langle x_i^{(1)}, x_j^{(1)} \rangle}{{\sqrt{d}}||x^{(1)}_i||_2 \cdot ||x^{(1)}_j||_2} 
        = \frac{\sqrt{d}(e^{\sqrt{d}}+j-1)}{\sqrt{e^{2\sqrt{d}} + i - 1} \sqrt{e^{2\sqrt{d}} + j - 1}}.
    \end{align}
    Since $i$ and $j$ are positive integers and the order of $j$ in the numerator is larger than that in the denominator, $S_{ij}^{(2)}$ is strictly increasing in $j$ for a fixed $i$.
\end{proof}

\subsection{Proof of Proposition \ref{nol2}}\label{proof_nol2}
\begin{proof}
Without LayerNorm,
\begin{align}
    S^{(1)}_{ij} = \mathrm{Causal}(X^{(0)}X^{(0)\top} / \sqrt{d})_{ij} = \begin{cases}
        \langle x_i, x_i \rangle / \sqrt{d} & (i=j) \\
        \langle x_i, x_j \rangle / \sqrt{d} & (i>j) \\
        -\inf & (i<j)
    \end{cases}.
\end{align}
Proceeding in the same manner as in the previous proof, we obtain
\begin{align}
    A^{(1)}_{ij} =
    \begin{cases}
        \frac{e^{\langle x_i, x_i \rangle / \sqrt{d}}}{e^{\langle x_i, x_i \rangle / \sqrt{d}}  + \sum_{k=1}^{i-1} e^{\langle x_i, x_k\rangle / \sqrt{d}} }& (i=j) \\
        \frac{e^{\langle x_i, x_j\rangle / \sqrt{d}}}{e^{\langle x_i, x_i \rangle / \sqrt{d}} + \sum_{k=1}^{i-1} e^{\langle x_i, x_k\rangle / \sqrt{d}} } & (i>j) \\
        0 & (i<j)
    \end{cases}.
\end{align}
Here, we can apply the law of large numbers. Thus,
\begin{align}
    \langle x^{(0)}_i, x^{(0)}_j\rangle = \begin{cases}
        1 & (i=j)\\
        \approx 0 & (i \neq j).
    \end{cases}
\end{align}
Therefore,
\begin{align}
    A_{ij}^{(1)} =
    \begin{cases}
        \frac{e^{1/\sqrt{d}}}{e^{1/\sqrt{d}} + (i-1)} & (i=j) \\
        \frac{1}{e^{1/\sqrt{d}} + (i-1)} & (i>j) \\
        0 & (i<j)
    \end{cases}.
\end{align}
Thus,
\begin{align}
    x_i^{(1)} = \sum_{j=1}^{i} A_{ij}^{(1)}x_j^{(1)} = \frac{e^{1/\sqrt{d}}x_i^{(1)}+\sum_{k=1}^{i-1}x_k^{(1)}}{e^{1/\sqrt{d}}+i-1},
\end{align}
\begin{align}
    \big(S_{ij}^{(2)}\big)_{i>j} &= \langle x_i^{(1)}, x_j^{(1)} \rangle = \frac{(e^{1/\sqrt{d}}x_i^{(1)}+\sum_{k=1}^{i-1}x_k^{(1)})(e^{1/\sqrt{d}}x_j^{(1)}+\sum_{l=1}^{j-1}x_l^{(1)})}{(e^{1/\sqrt{d}}+i-1)(e^{1/\sqrt{d}}+j-1)}  \\
    =& \frac{e^{1/\sqrt{d}} + j-1}
    {(e^{1/\sqrt{d}}+i-1)(e^{1/\sqrt{d}}+j-1)} \\
    =& \frac{1}{e^{1/\sqrt{d}}+i-1}.
\end{align}
Therefore, the attention score in the second layer without LayerNorm does not exhibit recency bias.\end{proof}

\subsection{Proof of Proposition \ref{residual} and \ref{anisotropic}}\label{apd:anisotropic}
Here, we consider both anisotropic input embeddings and residual connections.
To incorporate these effects, we rewrite Eq. \ref{eq:simplified} to include residual networks:
\begin{align}
        f(X) = \mathrm{Softmax}(\mathrm{Causal}(YY^\top) / \sqrt{d})Y + \gamma X,\\
        Y=\sqrt{d} X /||X||_2,
\end{align}
where $\gamma$ is a constant: $\gamma = 0$ corresponds to the model without residual connections, and $\gamma = 1$ corresponds to the model with residual connections.

\begin{proof}
We first compute the expectation of $\langle y_i^{(0)}, y_j^{(0)} \rangle$ when $i \neq j$.
For the denominator,
\begin{align}
    \mathbb{E}\big[||x_i^{(1)}||_2^2\big] &= \mathbb{E}\big[(\epsilon_i + \sqrt{\frac{\alpha}{1 - \alpha}}v)(\epsilon_i + \sqrt{\frac{\alpha}{1 - \alpha}}v)\big]\\
    &= \mathbb{E}[\langle \epsilon_i, \epsilon_i\rangle] + 2 \sqrt{\frac{\alpha}{1-\alpha}} \mathbb{E}[\langle \epsilon_i, v\rangle] + \frac{\alpha}{1-\alpha} \mathbb{E}[\langle v, v \rangle].
\end{align}
Since $d \gg 1$, the inner product in the second term is negligible compared to those in the first and third terms.
Thus, $\mathbb{E}\big[\||x_i^{(1)}||_2^2\big] \approx {\sqrt{d}}/({1-\alpha})$.
Therefore, 
\begin{align}
    \langle y_i^{(0)}, y_j^{(0)} \rangle = \sqrt{d}\frac{(\epsilon_i +cv) (\epsilon_j +cv)}{\frac{d}{1-\alpha}} = \sqrt{d}\alpha
\end{align}
Following the same procedure as in the above proof, we obtain
\begin{align}
    x_i^{(1)} &= \sum_{j=1}^{i} A_{ij}^{(1)}y_j^{(1)} + \gamma x_j^{(1)}\\
    &= \frac{e^{\sqrt{d}}y_i^{(1)}+ e^{\sqrt{d}\alpha}\sum_{k=1}^{i-1}y_k^{(1)}}{e^{\sqrt{d}} + e^{\sqrt{d}\alpha}(i-1)} + \sqrt{d}\gamma ||x^{(1)}_i||_2y_i^{(1)}.
\end{align}
We denote the denominator by $D_i$, so that
\begin{align}
    x_i^{(1)}= \frac{(d\gamma D_i+e^{\sqrt{d}})y_i^{(1)}+ e^{\sqrt{d}\alpha}\sum_{k=1}^{i-1}y_k^{(1)}}{D_i}
\end{align}

\begin{align}
    \big( \langle x_i^{(1)}, x_j^{(1)} \rangle \big)_{i>j} &= \frac{1}{D_i D_j} \bigg[ ((d\gamma D_i+e^{\sqrt{d}})y_i^{(1)}+ e^{\sqrt{d}\alpha}\sum_{k=1}^{i-1}y_k^{(1)}) \notag \\
    &\quad \times ((d\gamma D_j+e^{\sqrt{d}})y_j^{(1)}+ e^{\sqrt{d}\alpha}\sum_{l=1}^{j-1}y_l^{(1)}) \bigg] \\
    &= \frac{d}{D_i D_j} \bigg[ \alpha(d\gamma D_i+e^{\sqrt{d}})(d\gamma D_j+e^{\sqrt{d}}) + \alpha e^{\sqrt{d}\alpha}( d\gamma D_i+e^{\sqrt{d}})(j-1) \notag \\
    &\quad + \alpha e^{\sqrt{d}\alpha} (d\gamma D_j+e^{\sqrt{d}})(i-2) + e^{\sqrt{d}\alpha} (d\gamma D_j+e^{\sqrt{d}}) + e^{2\sqrt{d}\alpha}(j-1) \notag \\
    &\quad + \alpha e^{2\sqrt{d}\alpha} (i-2)(j-1) \bigg] \\
    &= \frac{d}{D_i D_j} \bigg[(j-1)(\alpha e^{\sqrt{d}\alpha}(d\gamma D_i+e^{\sqrt{d}}) + e^{2\sqrt{d}\alpha} + \alpha e^{2\sqrt{d}\alpha}(i-2)) \notag \\
    &\quad + (d\gamma D_j+e^{\sqrt{d}})(\alpha (d \gamma D_i+e^{\sqrt{d}}) + \alpha e^{\sqrt{d}\alpha} (i-2) + e^{\sqrt{d}\alpha})\bigg] \\
    &= d\frac{(e^{\sqrt{d}\alpha}(j-1) + d\gamma D_j + e^{\sqrt{d}})(\alpha(d\gamma D_i + e^{\sqrt{d}}) + e^{\sqrt{d}\alpha} + \alpha e^{\sqrt{d}\alpha} (i-2)) }{D_i D_j} \\
    &= d\frac {D_j (1 +d\gamma) (\alpha(d\gamma D_i + e^{\sqrt{d}}) + e^{\sqrt{d}\alpha} + \alpha e^{\sqrt{d}\alpha} (i-2))}{D_i D_j} \\
    &= d\frac {(1 +d\gamma) (\alpha(d\gamma D_i + e^{\sqrt{d}}) + e^{\sqrt{d}\alpha} + \alpha e^{\sqrt{d}\alpha} (i-2))}{D_i}
\end{align}
Formally, we aim to show that $S_{ij}^{(2)} > S_{ik}^{(2)}$ for any $i \geq j > k$.
Since $S_{ij}^{(2)} = d\frac{\langle x_i^{(1)}, x_j^{(1)} \rangle}{||x_i^{(1)}|| \cdot ||x_j^{(1)}||}$, and the numerator and $||x_i^{(1)}||_2$ are independent of $j$, it suffices to show that $||x_j^{(1)}||_2$ is strictly decreasing in $j$.

\begin{align}
    ||x_j^{(1)}||^2_2 &= \frac{((\gamma D_j+e^{\sqrt{d}})y_j^{(1)}+ e^{\alpha\sqrt{d}}\sum_{k=1}^{j-1}y_k^{(1)})((\gamma D_j+e^{\sqrt{d}})y_j^{(1)}+ e^{\alpha\sqrt{d}}\sum_{k=1}^{j-1}y_k^{(1)})}{D_j D_j} \\
    &= \frac{d}{D_j^2} \bigg[(\gamma D_j+e^{\sqrt{d}})^2 + 2d\alpha(\gamma D_j+e^{\sqrt{d}}) de^{\alpha\sqrt{d}}(j-1) + e^{2\alpha\sqrt{d}} (j-1) \nonumber \\
    &\quad + d\alpha e^{2\alpha\sqrt{d}} (j-2)(j-1) \bigg] \\
    &= \frac{d}{D_j^2}\bigg[(\gamma D_j+e^{\sqrt{d}})^2 + 2(\gamma D_j+e^{\sqrt{d}})\alpha (D_j - e^{\sqrt{d}}) + e^{\sqrt{d}\alpha}(D_j - e^{\sqrt{d}}) \nonumber \\
    &\quad+ \alpha (D_j - e^{\sqrt{d}} -e^{\sqrt{d}\alpha})(D_j - e^{\sqrt{d}})\bigg] \\
    &= d\frac{D_j^2 (\gamma ^2 + 2\alpha \gamma + \alpha) + D_j (1-\alpha)(2\gamma e^{\sqrt{d}} + e^{\alpha\sqrt{d}}) + (1-\alpha) (e^{2\sqrt{d}} - e^{\sqrt{d} + \sqrt{d}\alpha})}{D_j^2} 
\end{align}
Since $||x_j^{(1)}||_2 \geq 0$, it suffices to show that $||x_{j+1}^{(1)}||_2^2 - ||x_j^{(1)}||_2^2 < 0$:
\begin{align}
    &||x_{j+1}^{(1)}||^2_2 - ||x_j^{(1)}||^2_2  \\
    &= d(1-\alpha)(2\gamma e^{\sqrt{d}} + e^{\alpha\sqrt{d}})({1 \over D_{j+1}} - {1 \over D_j}) + d(1-\alpha) (e^{2\sqrt{d}} - e^{\sqrt{d}+\alpha\sqrt{d}})({1 \over D_{j+1}^2} - {1 \over D_j^{2}}) \\
    &= d(1-\alpha)(2\gamma e^{\sqrt{d}} + e^{\alpha\sqrt{d}})({-e^{\alpha\sqrt{d}} \over D_j D_{j+1}}) + d(1-\alpha) (e^{2\sqrt{d}} - e^{\sqrt{d}+\sqrt{d}\alpha})({-e^{\alpha\sqrt{d}} (2D_j + e^{\alpha\sqrt{d}}) \over D_j^2 D_{j+1}^2})
\end{align}
Since $0 \leq \alpha < 1$, both terms are strictly negative.
Therefore, $||x_j^{(1)}||_2$ is strictly decreasing in $j$.
Consequently, $S_{ij}^{(2)}$ is strictly increasing in $j$ for fixed $i$, exhibiting a recency bias regardless of the presence of residual connections or the anisotropy of the input embeddings.
\end{proof}

From the above proof, since $\langle x_i^{(1)}, x_j^{(1)} \rangle$ is independent of $j$, we have
\begin{align}
    \big( S_{ij}^{(2)} \big)_{i>j} = d\frac{\langle x_i^{(1)}, x_j^{(1)} \rangle}{||x_i^{(1)}|| \cdot ||x_j^{(1)}||} = \bigg(d \frac{\langle x_i^{(1)}, x_j^{(1)} \rangle}{||x_i^{(1)}||} \bigg)\bigg(\frac{1}{||x_j^{(1)}||}\bigg) = g(i) \cdot h(j).
\end{align}
Therefore,
\begin{align}
    h(j) = \frac{D_j}{\sqrt{d}\sqrt{D_j^2 (\gamma ^2 + 2\alpha \gamma + \alpha) + D_j (1-\alpha)(2\gamma e^{\sqrt{d}} + e^{\alpha\sqrt{d}}) + (1-\alpha) (e^{2\sqrt{d}} - e^{\sqrt{d} + \sqrt{d}\alpha})}}.\label{eq:hj}
\end{align}

\newpage
\section{Extended Results}\label{apd:res}

\begin{figure}[!h]
    \centering
    \includegraphics[width=\linewidth]{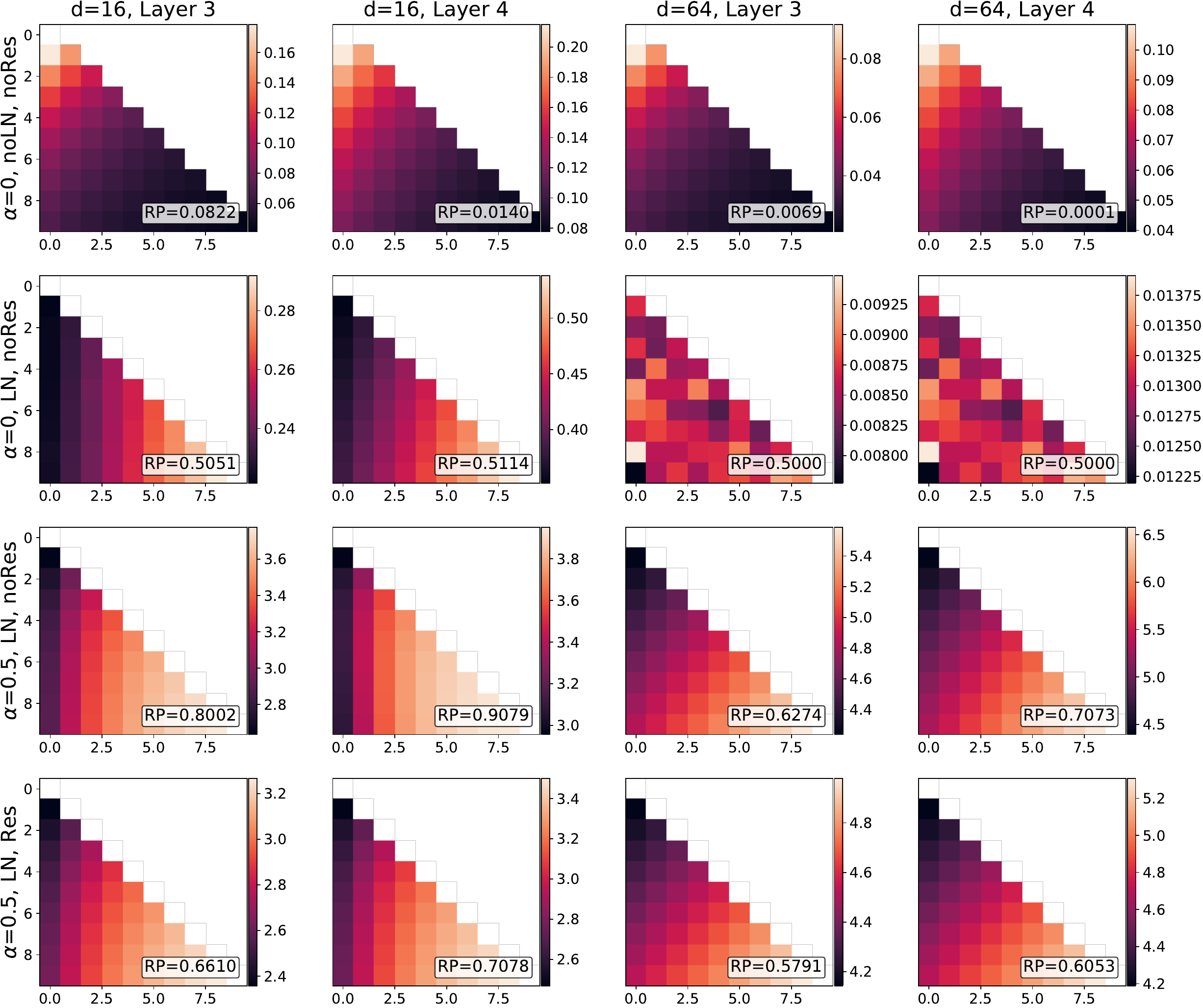}
    \caption{Visualization of the attention scores for layers 3 and 4 with $d=16$ (left) and $d=64$ (right). The layout follows Figure~\ref{fig:l2norm}.}
    \label{fig:extended_layers}
\end{figure}

\begin{figure}[!h]
    \centering
    \includegraphics[width=\linewidth]{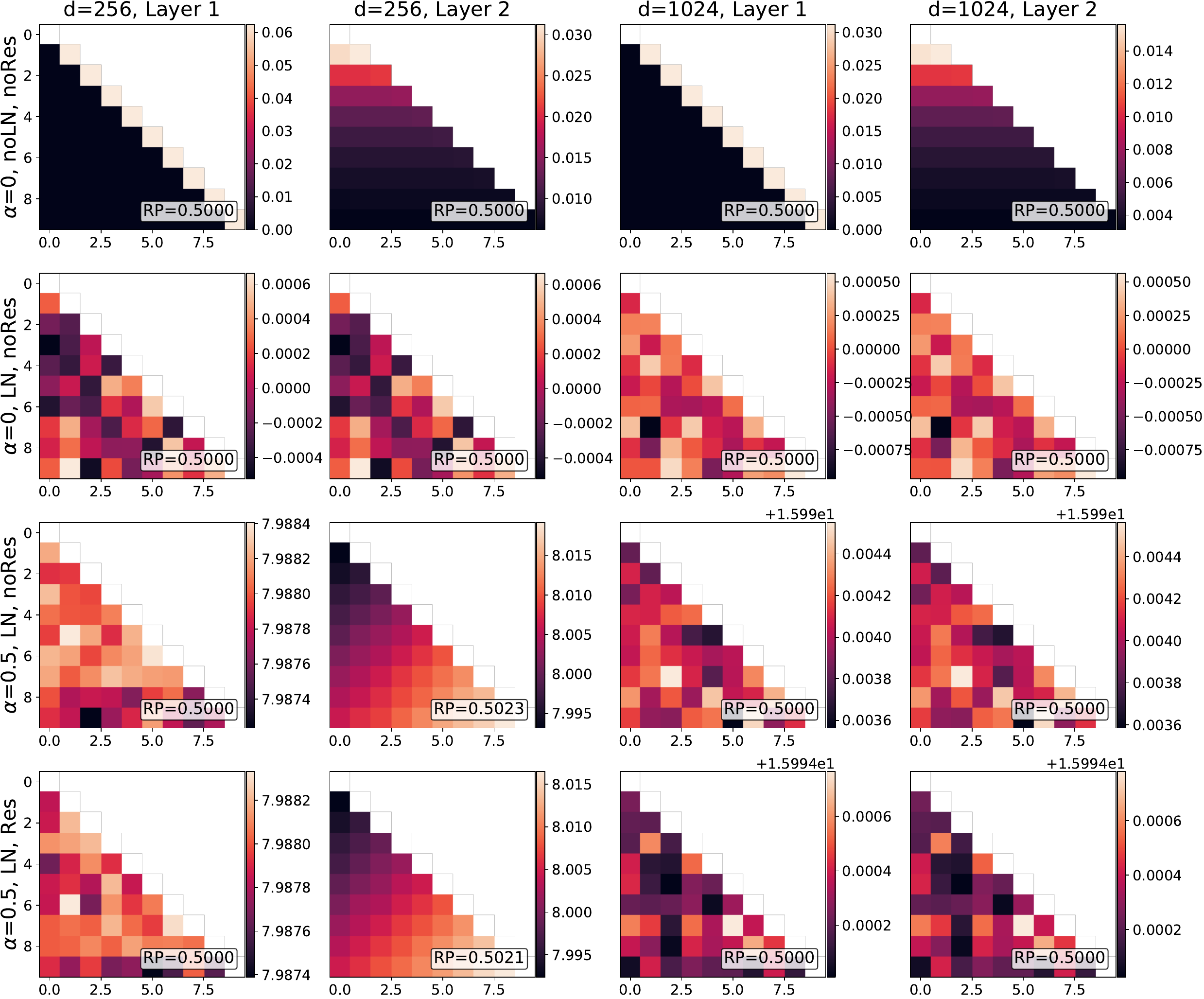}
    \caption{Visualization of the attention scores for layers 1 and 2 with $d=256$ (left) and $d=1024$ (right). The layout follows Figure~\ref{fig:l2norm}.}
    \label{fig:extended_dims}
\end{figure}

\begin{figure}[!h]
    \centering
    \includegraphics[width=\linewidth]{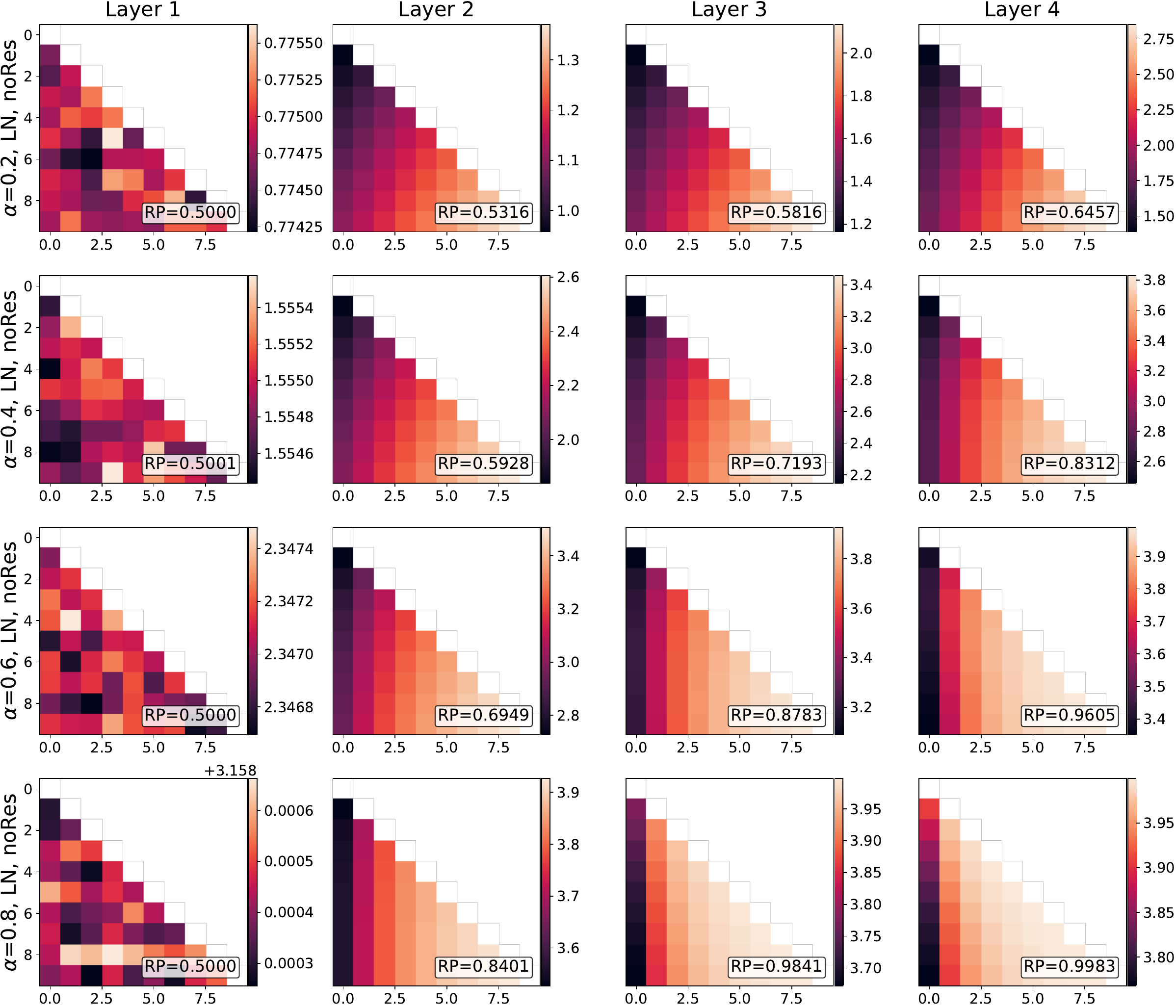}
    \caption{Extended results of Figure \ref{fig:l2norm} with multiple $\alpha$ values and no residual connections.}
    \label{fig:extended_alpha}
\end{figure}

\begin{figure}
    \centering
    \includegraphics[width=\linewidth]{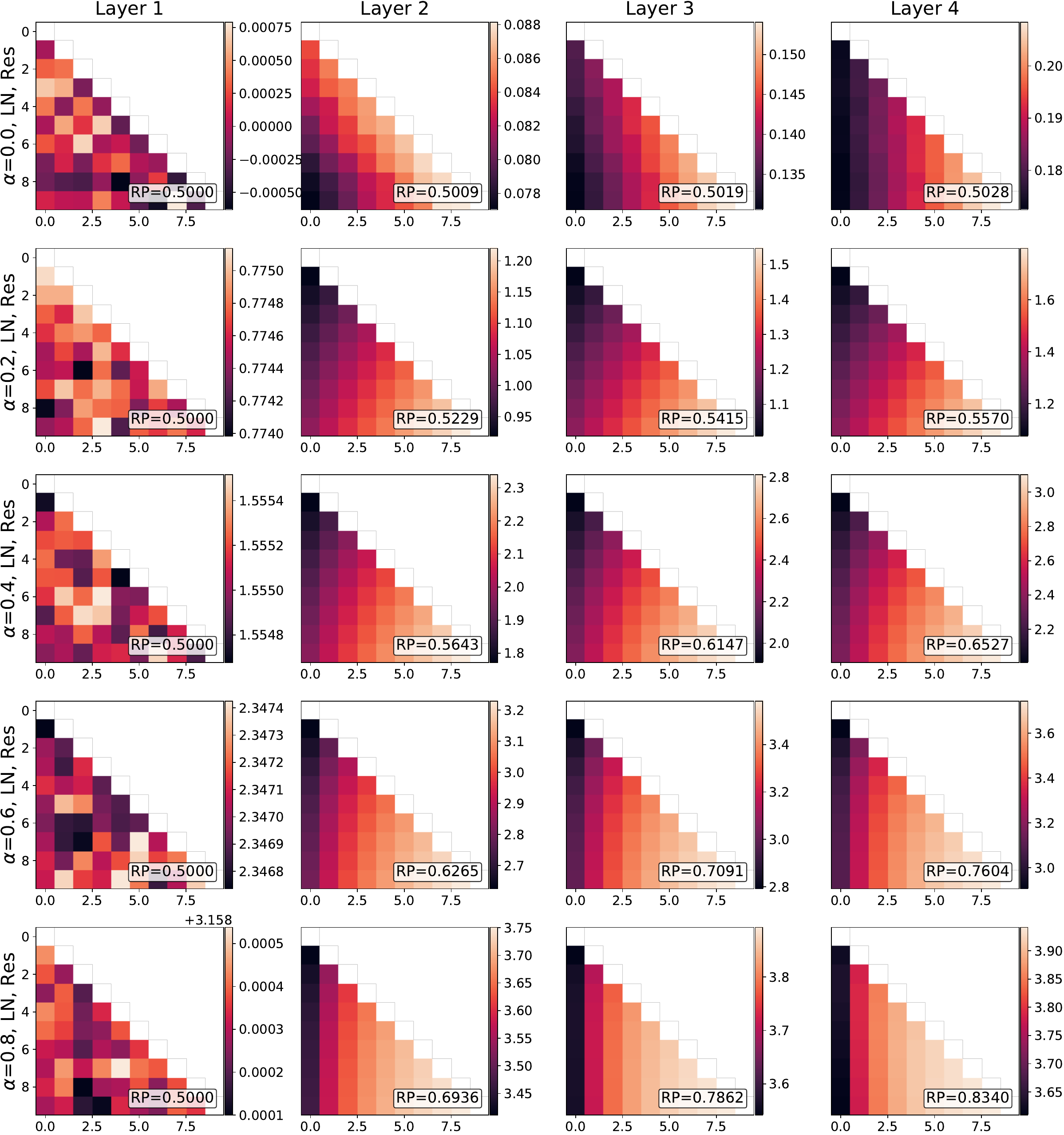}
    \caption{Extended results of Figure \ref{fig:l2norm} with multiple $\alpha$ values and with residual connections.}
    \label{fig:extended_alpha_resid}
\end{figure}

\begin{figure}
    \centering
    \includegraphics[width=\linewidth]{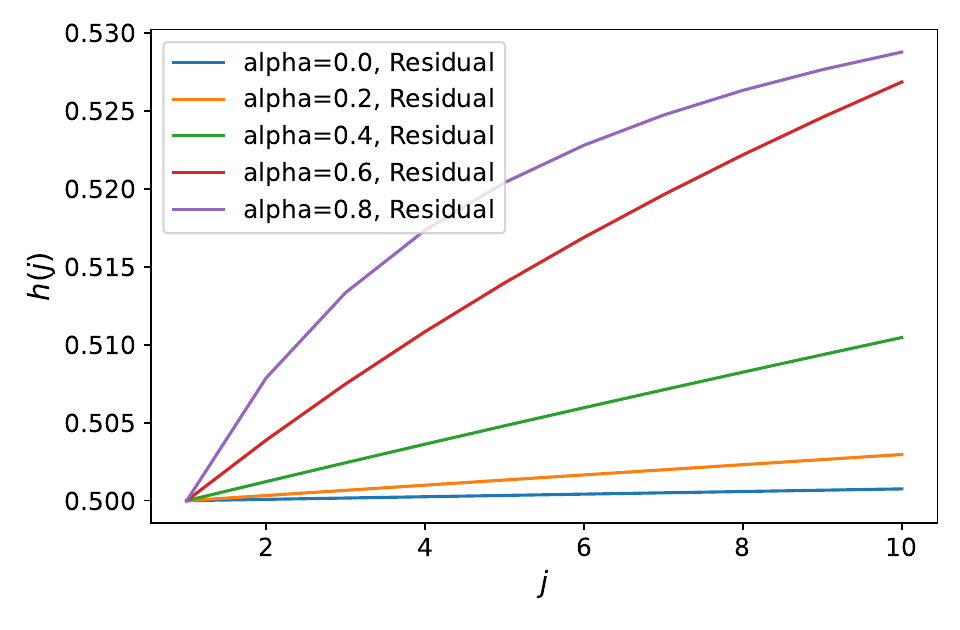}
    \caption{Visualization of $h(j)$ over key index $j$, for multiple values of $\alpha$, including residual connections.}
    \label{fig:alpha_resid}
\end{figure}

\end{document}